\documentclass[runningheads]{llncs}
\usepackage[T1]{fontenc}
\usepackage{graphicx}
\usepackage{latexsym}

\usepackage[T1]{fontenc}

\usepackage[utf8]{inputenc}

\usepackage{microtype}

\usepackage{inconsolata}

\usepackage{graphicx}
\usepackage{tabularx}
\usepackage{threeparttable} 

\usepackage{multirow}
\usepackage{amsmath, amssymb}
\usepackage{enumitem}
\usepackage{amsmath}
\usepackage{float}
\usepackage{booktabs}

\usepackage{algorithm}
\usepackage{algpseudocode}

\usepackage{multirow}
\usepackage{siunitx}
\usepackage{graphicx}   
\usepackage{caption}    
\usepackage{subcaption} 
\usepackage{xcolor}     
\usepackage{float}      
\usepackage{hyperref}

\definecolor{darkgreen}{RGB}{0,100,0} 
\hypersetup{
    colorlinks=true,
    urlcolor=blue,
    linkcolor=blue,
    citecolor=blue
}
\DeclareMathOperator*{\argmax}{arg\,max}

\usepackage{color}

\newcommand{\modelxcom}{{\fontfamily{cmss}\selectfont XCom}}
\newcommand{\gitrepo}{\url{https://anonymous.4open.science/r/XCom}}

\begin{document}
\title{XAI-enhanced Comparative Opinion Mining \\via Aspect-based Scoring and Semantic Reasoning}
%
%



\author{Ngoc-Quang Le\inst{1}\thanks{These authors contributed equally to this work.} \and
T. Thanh-Lam Nguyen\inst{1}\protect\footnotemark[1] \and
Quoc-Trung Phu\inst{1} \and
Thi-Phuong Le\inst{1,2} \and
Duy-Cat Can\inst{1,3,4}\orcidID{0000-0002-6861-2893} \and
Hoang-Quynh Le\inst{1}\orcidID{0000-0002-1778-0600}}

\authorrunning{Le et al.}
%

\institute{VNU University of Engineering and Technology, 144 Xuan Thuy, Cau Giay, Hanoi, Vietnam\
\email{{22024510,22024516,21020096,phuong.lt,catcd,lhquynh}@vnu.edu.vn}\
\url{https://uet.vnu.edu.vn} \and
School of Computing and Information Systems, Singapore Management University, 81 Victoria Street, Singapore 188065, Singapore\
\email{tp.le.2023@phdcs.smu.edu.sg}\
\url{https://scis.smu.edu.sg} \and
Centre Hospitalier Universitaire Vaudois, Rue du Bugnon 46, 1011 Lausanne, Switzerland\
\email{duy-cat.can@chuv.ch} \and
University of Lausanne, Quartier UNIL-Centre, 1015 Lausanne, Switzerland\
\email{duy-cat.can@chuv.ch}\
\url{https://www.unil.ch}}
\titlerunning{XAI-enhanced Comparative Opinion Mining}
\maketitle              

\begin{abstract}
Comparative opinion mining involves comparing products from different reviews. However, transformer-based models designed for this task often lack transparency, which can adversely hinder the development of trust in users. In this paper, we propose \modelxcom{}, an enhanced transformer-based model separated into two principal modules, i.e., \textit{(i)} aspect-based rating prediction and \textit{(ii)} semantic analysis for comparative opinion mining. \modelxcom{} also incorporates a Shapley additive explanations module to provide interpretable insights into the model’s deliberative decisions. 
Empirically, \modelxcom{} achieves leading performances compared to other baselines, which demonstrates its effectiveness in providing meaningful explanations, making it a more reliable tool for comparative opinion mining.
Source code is available at: \gitrepo{}.

\end{abstract}

\section{Introduction}

\begin{figure}[ht!]
  \centering
  \includegraphics[width=\linewidth]{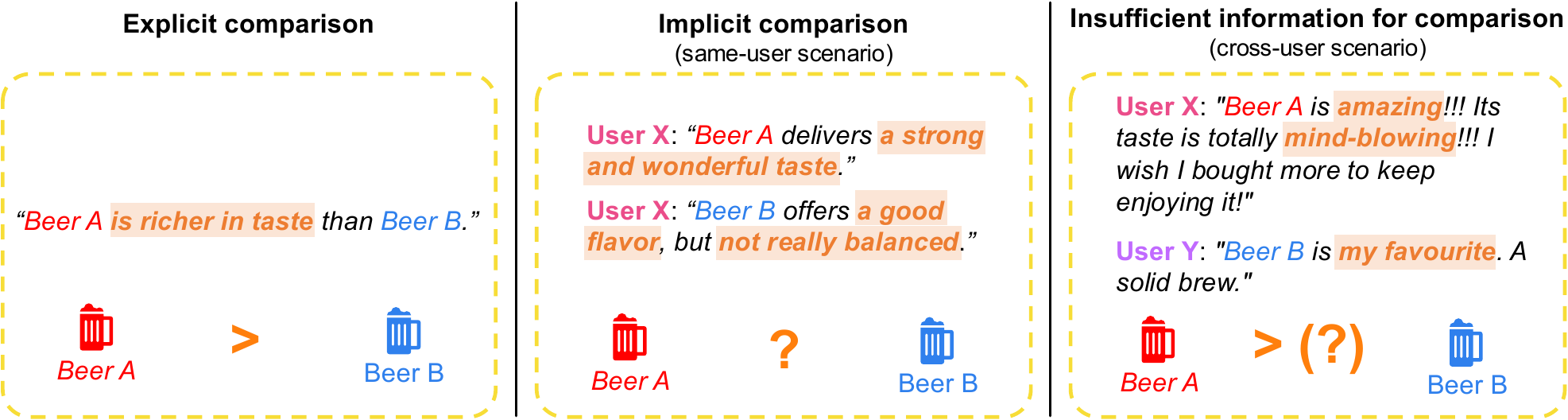}
  \caption{\small
  Illustration of explicit vs. implicit comparative opinions in user reviews. 
  \textbf{Explicit:} Direct comparison using terms like ``\textit{is richer in taste than}''. 
  \textbf{Implicit:} Two separate reviews are from the same user. Preference is implied, and not stated, which is harder to detect.
  \textbf{Insufficient information in the cross-user setting:} Two reviews are from different users. Comparisons are unreliable due to subjective differences in user standards (e.g., leniency, writing style), lacking a common baseline for inference.}
  \label{fig:implicit_comparison}

\end{figure}

In the contemporary marketplace, choice overload may occur when users are often presented with an excessive number of options. Reviews tend to convey personal experiences rather than provide direct comparisons across alternatives. 
As illustrated in Figure~\ref{fig:implicit_comparison}, explicit comparisons clearly indicate preferences between products, whereas implicit ones are often distributed across multiple reviews, requiring contextual reasoning to infer user intentions. Although such reviews provide valuable information, the lack of structured comparative content hinders customers from identifying which aspects are most influential or which product performs better overall. 
Comparative opinion mining bridges this gap by systematically analyzing and contrasting different options~\cite{varathan2017comparative}. It helps users discern key distinctions and make more informed decisions.

The same-user setting is also our core objective to mitigate the noise arising from individual behavioral variances, specifically linguistic style, rating leniency, and personalized aspect weighting. Consider a ``strict'' user who typically gives low ratings and writes brief, reserved summaries, versus a ``lenient'' user known for high ratings and detailed reviews. Paradoxically, a rating of ``\textit{good}'' from the former often implies higher quality than ``\textit{amazing}'' from the latter. By isolating comparisons to the same user, we remove these inconsistencies and capture preferences against a stable, personalized baseline. 

The increasing use of advanced deep learning techniques and large language models holds substantial promise for enhancing performance in comparative opinion mining tasks. However, a significant challenge persists as most of these models operate as black boxes, producing comparative conclusions without providing underlying reasoning \cite{minh2022explainable}. This lack of transparency can diminish users' trust due to the difficulty in understanding the rationale behind a particular decision \cite{mersha2024explainable}. 
An unexplained comparative statement may offer little value -- especially when users are making fine-grained decisions based on specific product attributes. Therefore, it is essential to incorporate an explanation ability that provides meaningful reasoning to support the model's decisions.

To address these challenges in Comparative Opinion Mining, we present \modelxcom{}, 
\textbf{X}AI-enhanced \textbf{C}omparative \textbf{O}pinion \textbf{M}ining. \modelxcom{} is a Transformer-based model that integrates aspect-based rating prediction techniques and a semantic analysis framework to address the problem of comparing two reviews across multiple aspects. 
Additionally, we enhance the model's reliability by integrating an explainability phase inspired by SHAP (SHapley Additive exPlanations) \cite{lundberg2017unified}, which highlights the key factors influencing the model's decision-making process. 


\section{Related Work}

Jindal and Liu (2006) \cite{jindal2006mining} laid the foundational groundwork for mining comparative sentences and relations, initiating research on identifying and analyzing comparative expressions to assess product performance. Le and Lauw (2021) \cite{le2021explainable} later explored the generation of comparative sentences using templates or existing comparative expressions.
By synthesizing comparative data, Shi et al. (2023) \cite{shi2023ratingreview} and Hasan et al. (2025) \cite{hasan2025reviewbased} identify which product performs better based on user-generated content, providing insights into consumer preferences and product rankings for recommendation systems and market analysis. 

In parallel, Gao et al. (2024) \cite{gao2024end} and Le et al. (2024) \cite{le2024overview} focus on extracting information from comparative sentences. Their approaches often involve identifying specific features (e.g., price, quality) compared across products, which Zhao et al. (2024) \cite{zhao2024enhancing} have shown to be particularly valuable for e-commerce applications. These approaches, while effective in controlled settings, limit flexibility and scalability across product domains. Similarly, Echterhoff et al. (2023) \cite{echterhoff2023comparing} generate comparative sentences for individual products but cannot handle arbitrary product pairs, restricting their use in dynamic real-world scenarios.

Despite these advancements, existing studies face several critical limitations such as reliance on comparative sentences, difficulty with arbitrary product pairs, and lack of explainability. 
Our framework alleviates such problems by digging deeper into implicit pairwise comparative opinions and leveraging SHAP for interpretability. 
Moreover, while prior works have explored explaining deep learning (Selvaraju et al. (2017) \cite{selvaraju2017grad}) and Transformer-based models (Abnar and Zuidema (2020) \cite{abnar2020quantifying} and Song et al. (2024) \cite{song2024better}), applying such methods to complex architectures remains challenging. 
SHAP paves the way for this problem to be tackled by analyzing token contributions through the model's output changes. 
Therefore, it can be applied to any model, regardless of internal complexity.

\section{Methods}

\begin{figure}[h!]
  \centering
  \includegraphics[width=0.8\columnwidth]{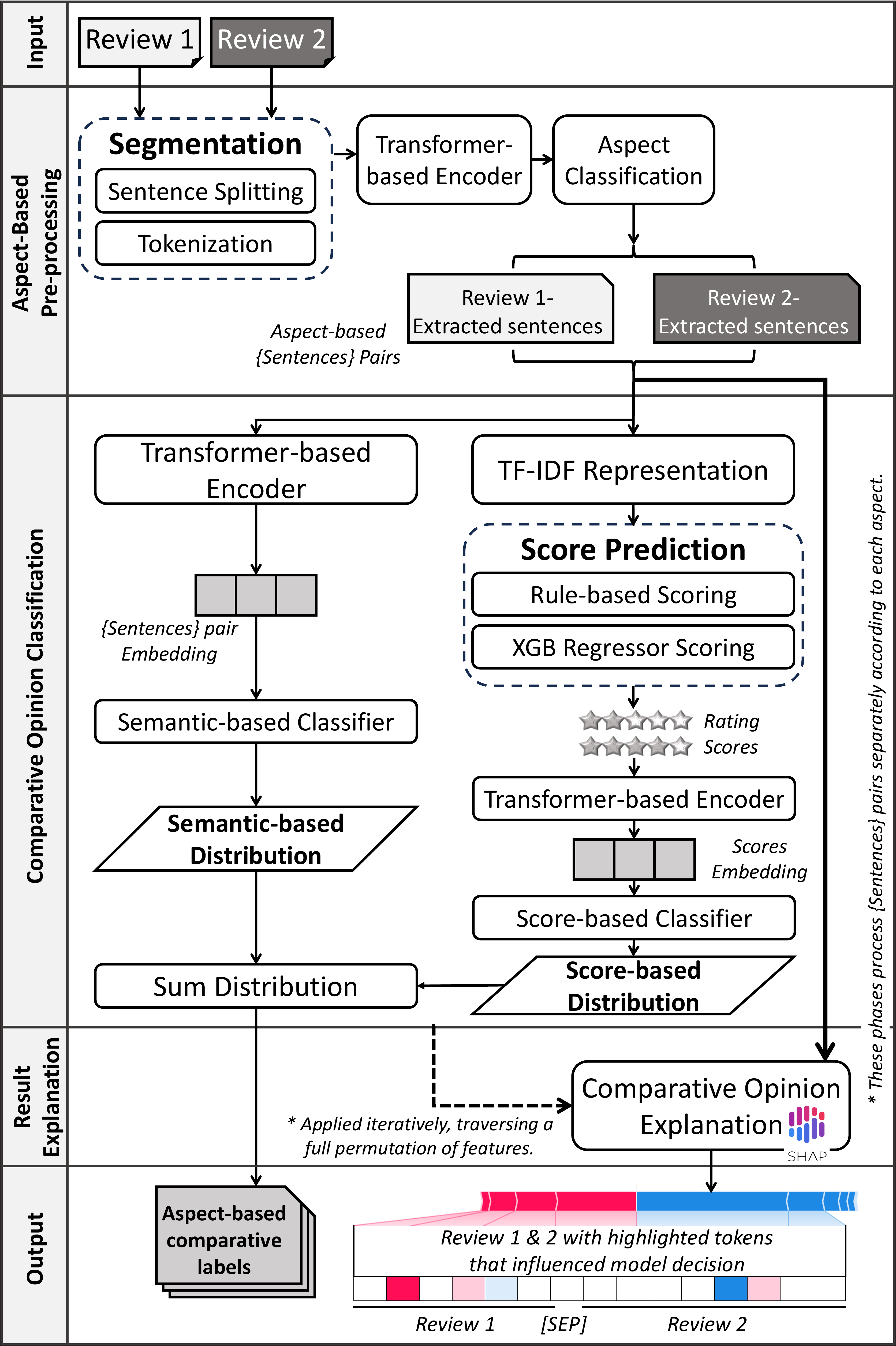}
  \caption{\small \modelxcom{} architecture.}
  \label{fig:model}

\end{figure}

Figure~\ref{fig:model} illustrates the proposed model, comprising three main phases:
Aspect-Based Pre-processing,
Comparative Opinion Classification,
Prediction \& Explanation.

\subsection{Aspect-Based Pre-processing}

The aspect-based pre-processing phase converts a review pair into structured aspect-specific sentences pairs. It applies normalization, sentence splitting, tokenization, and aspect classification to transform raw reviews for further analysis.

Given a user set $\mathcal{U}$ and a dataset $\mathcal{D} = \{(R_1^u, R_2^u) \mid u \in \mathcal{U} \}$,
each comparative instance $(R_1^u, R_2^u)$ consists of two reviews selected from the same user $u$. 
Each review $R_i^u$ contains sentences $R_i^u = [ s_1^i, s_2^i, \dots, s_{N_i}^i ]$,
%
where $s_j^i$ is the $j$-th sentence in review $i$, and $N_i$ is the total number of sentences. 

We use four BERT-based classifiers $f_a$, each trained for a distinct aspect $a \in \mathcal{A}$.
Each classifier assigns $1$ if a sentence corresponds to aspect $a$, otherwise $0$.
The classified sentences are then grouped for comparison $P_a^u = (s_a^1, s_a^2)$,
where $s_a^1$ and $s_a^2$ are the extracted sentences from the first and second reviews for aspect $a$.
The final output is a structured set of aspect-based sentence tuples:
\begin{equation}
    \mathcal{P}^u = \{ P_a^u \mid a \in \mathcal{A} \}.
\end{equation}

\subsection{Comparative Opinion Classification}
\paragraph{Score-based classifier:}

Following the aspect-based pre-processing phase, each review pair is structured into four aspect-specific tuples.
These sentence pairs are used to compute aspect-wise rating scores.

For a given aspect $a \in \mathcal{A}$, extracted sentences from reviews $R_1^u$ and $R_2^u$ form two sets: $S_a^1$ and $S_a^2$.
Each sentence's \textbf{rating score} is determined by averaging the sentiment scores of its adjectives: 
\begin{equation}
    \text{rating score}\left( s_i^k \right) = \frac{1}{N_i^k} \sum_{j=1}^{N_i^k} \text{score} \left( w_j \right),
\end{equation}
where $N_i^k$ is the number of adjectives in sentence $s_i^k$, and $\text{score} \left( w_j \right)$ is the sentiment score of adjective $w_j$.  
If $N_i^k = 0$, meaning the sentence $s_i^k$ contains no adjectives from the predefined sentiment dictionary, we compute its TF-IDF feature vector. This vector is then fed into an aspect-specific XGBoost regressor, which has been trained to predict rating scores in cases where direct adjective-based scoring is not possible. The predicted rating score is given by:
%
\begin{equation}
    \text{rating score}\left( s_i^k \right) = f^{\texttt{XGB}}\left( \text{TF-IDF}\left( s_i^k \right) \right),
\end{equation}
where $f^{\texttt{XGB}}$ maps TF-IDF vectors to rating scores.
Since each aspect contains multiple sentences, the overall aspect rating for a review is determined by the minimum sentence rating:
%
\begin{equation}
    \text{rating score}\left( S_a^k \right) = \min_{i \in \left\{1, \dots, M_a^k \right\}} \text{rating score}\left( s_i^k \right),
\end{equation}
%
where $M_a^k$ is the number of sentences in $S_a^k$.

For \textbf{classification}, aspect rating scores from both reviews are concatenated into a structured input:
$\texttt{input}^\texttt{r} = [\texttt{CLS}] \ \text{rating score}(S_a^1) \ [\texttt{SEP}] \ \text{rating score}(S_a^2) \ [\texttt{SEP}]$.
%
where $[\texttt{CLS}]$ is the classification token, and $[\texttt{SEP}]$ is the separator.  
%
A Transformer-based encoder extracts rating score embeddings:  
%
$\mathbf{z}^\texttt{r} = \mathcal{E}^\texttt{r}(\texttt{input}^\texttt{r})\mid_{\texttt{CLS}}$,
where $\mathbf{z}^\texttt{r} \in \mathbb{R}^{768}$ is the \texttt{CLS} token’s logits vector. 
%
Finally, a score-based classifier processes $\mathbf{z}^\texttt{r}$ to produce a distribution over three comparative classes:  
\begin{equation}
    \mathbf{l}^\texttt{r} = \mathbf{z}^\texttt{r} \cdot \mathbf{W}^\texttt{r} + \mathbf{b}^\texttt{r},
\end{equation}
%
where $\mathbf{W}^\texttt{r} \in \mathbb{R}^{768 \times 3}$ is the weight matrix, and $\mathbf{b}^\texttt{r} \in \mathbb{R}^{3}$ is the bias vector.

\paragraph{Semantic-based classifier:}
This step directly processes sentence pairs to generate embeddings, capturing semantic relationships.
The structured input is:  
$\texttt{input}^\texttt{s} = [\texttt{CLS}] \ (S_a^1) \ [\texttt{SEP}] \ (S_a^2) \ [\texttt{SEP}] \ [\texttt{PAD}]$.
%
A Transformer-based encoder extracts sentence-pair embeddings:
%
$\mathbf{z}^\texttt{s} = \mathcal{E}^\texttt{s}(\texttt{input}^\texttt{s})\mid_{\texttt{CLS}}$,
where $\mathbf{z}^\texttt{s} \in \mathbb{R}^{768}$. 
%
The semantic-based classifier processes $\mathbf{z}^\texttt{s}$ to produce a score-based distribution:  
\begin{equation}
    \mathbf{l}^\texttt{s} = \mathbf{z}^\texttt{s} \cdot \mathbf{W}^\texttt{s} + \mathbf{b}^\texttt{s},
\end{equation}
%
where $\mathbf{W}^\texttt{s} \in \mathbb{R}^{768 \times 3}$ is the weight matrix, and $\mathbf{b}^\texttt{s} \in \mathbb{R}^{3}$ is the bias vector.

\paragraph{Comparative classification:}
To obtain the final probability distribution, we first apply $\text{softmax}$ to the score-based and semantic-based logits ($\mathbf{l}^\texttt{r}$ and $\mathbf{l}^\texttt{s}$), then sum the resulting probability vectors:
%
%
%
%
\begin{gather*}
    \textbf{p}^\texttt{r} = \text{softmax}(\mathbf{l}^\texttt{r}),
    \quad \textbf{p}^\texttt{s} = \text{softmax}(\mathbf{l}^\texttt{s}), \\
    \textbf{p}^\texttt{final} = \textbf{p}^\texttt{r} + \textbf{p}^\texttt{s}.
\end{gather*}
%
%
%
%
The predicted class $\hat{C}$ corresponds to the highest probability value:
\begin{equation}
    \hat{C} = \argmax_{k} \left( \textbf{p}^\texttt{final} \right)_k, \quad k \in \{1,2,3\}
\end{equation}

\subsection{SHAP-based Explanation}

We use SHAP~\cite{Shapley1951NotesOT} to enhance model transparency by attributing feature importance. It quantifies each feature's contribution by testing different subsets and observing their impact on predictions. SHAP provides model interpretability by revealing which tokens drive \modelxcom{}'s comparative predictions, thereby transforming it from a black-box into a white-box system. In other words, SHAP explanations serve to expose the model's internal decision attribution, not to validate human-aligned reasoning.

%
SHAP approximates the model $f$ with an interpretable linear function:
\begin{equation}
    g(z') = \phi_0 + \sum_{j=1}^{M} \phi_j z_j'\ ,
\end{equation} 
where $g(z')$ approximates $f(x)$, $\phi_j$ is the Shapley value of feature $j$, and $z' \in \{0,1\}^M$ indicates feature presence ($z_j' = 1$) or absence ($z_j' = 0$). The  prediction $\phi_0$ is the model’s average output.
Each feature's Shapley value is computed as:  
%
\begin{equation}
    \phi_i = \sum_{\mathcal{S} \subseteq \mathcal{N} \setminus \{i\}} \frac{|\mathcal{S}|!(|\mathcal{N}| \text{--} |\mathcal{S}| \text{--} 1)!}{|\mathcal{N}|!} \left[ f(\mathcal{S} \cup \{i\}) \text{--} f(\mathcal{S}) \right]
\end{equation}
where $\mathcal{N}$ is the set of all features, and $\mathcal{S}$ is a subset excluding feature $i$.
$f(\mathcal{S})$ is the model output with features in $\mathcal{S}$, while $f(\mathcal{S} \cup \{i\})$ includes $i$.  

In our case, where input consists of tokens, each token is treated as a feature. SHAP selects different subsets by replacing certain tokens with the \texttt{[MASK]} token, effectively removing them from the input. It then measures how the remaining tokens influence the model’s prediction.

\section{Experimental Results}

\subsection{Dataset}
Given the emerging and challenging nature of the task, along with the difficulty of obtaining highly specialized annotations, our experiments are conducted on a single dataset -- SUDO \cite{nguyen2026comparingsayingdatasetbenchmark}. In SUDO, each user provides multiple reviews, 
which are then paired to create comparative instances for implicit opinion mining (See \cite{nguyen2026comparingsayingdatasetbenchmark} for details on the dataset construction methodology).
Each review includes ratings for four aspects $\mathcal{A} = \{\text{appearance}, \text{aroma}, \text{palate}, \text{taste}\}$.
The dataset was annotated at two levels:
\begin{enumerate}[noitemsep,topsep=0pt]
    \item \textbf{Sentence-Aspect Level}: Each sentence is labeled as \texttt{1} (Mentioned) if it refers to a specific aspect or \texttt{0} (Not Mentioned) otherwise.
    
    \item \textbf{Review-Comparative Level}: Each review pair is labeled per aspect as: \texttt{-1} (worse -- the first review is rated lower than the second for that aspect), \texttt{0} (similar or non-comparable), \texttt{1} (better -- the first review is rated higher than the second), or \texttt{Null} (aspect not addressed).
\end{enumerate}

\begin{figure}[ht!]
  \centering
  \includegraphics[width=\columnwidth]{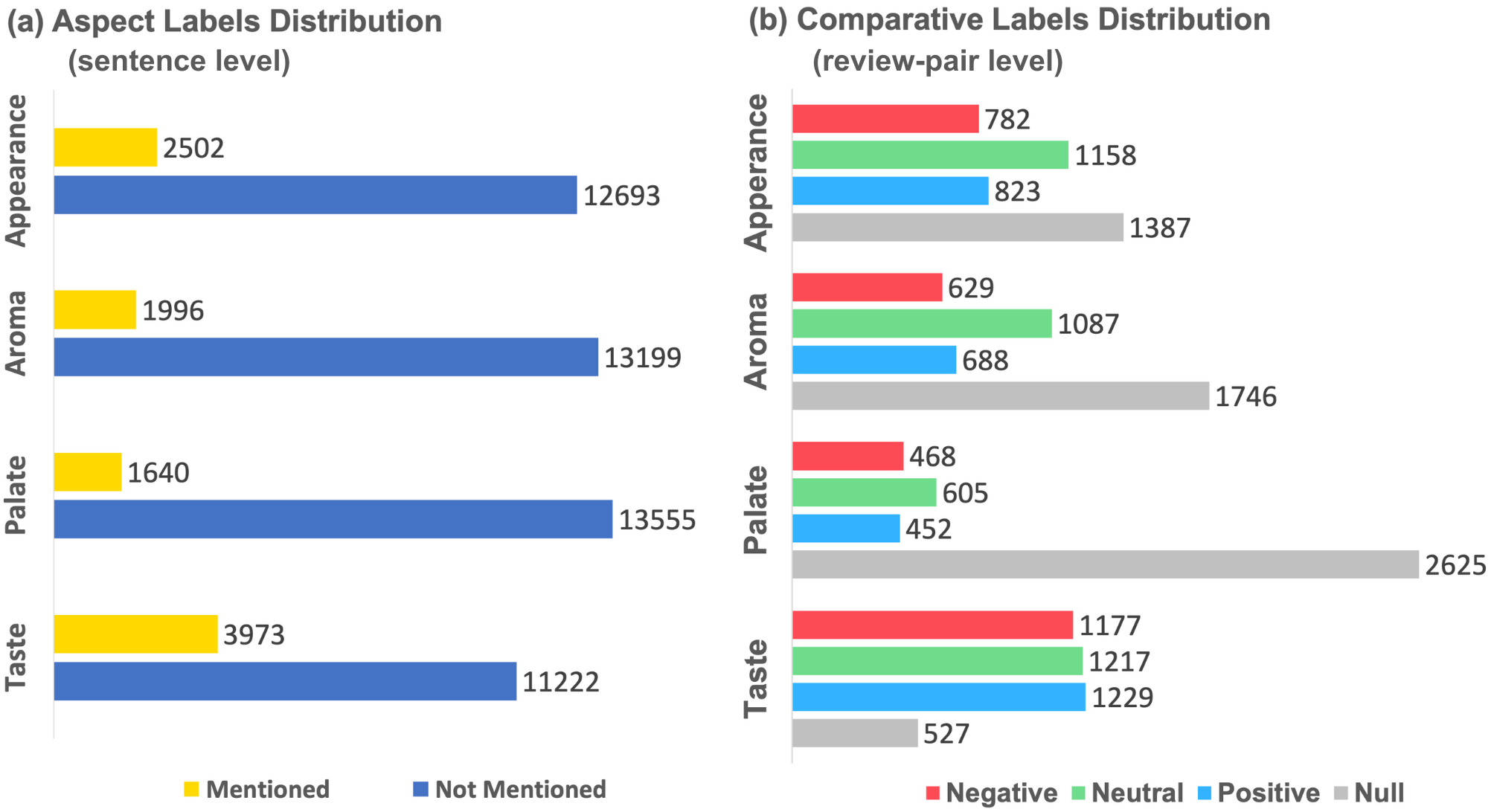}
  \caption{\small Distribution of aspect and comparative labels.}
  \label{fig:labels_distributions}

\end{figure}


At the sentence level, the class distribution across aspects in Figure~\ref{fig:labels_distributions} (a) reveals a significant imbalance. Sentences that do not mention an aspect vastly outnumber those that do. This underscores the challenge of accurately identifying relevant aspects within reviews, necessitating robust architectures to address it.


At the review level, Figure~\ref{fig:labels_distributions} (b) shows a high prevalence of the \texttt{Null} label across aspects: \texttt{taste} exhibits the lowest proportion, underscoring its central role in beer evaluation, whereas \texttt{palate} shows the highest, suggesting it is infrequently discussed. Among non-null pairs, the sentiment distribution reveals a slight tendency toward \texttt{Neutral}, indicating that when an aspect appears in both reviews, the associated sentiments are often consistent.
Overall, these significant concentrations pose a challenge for developing effective supervised models.

\subsection{Evaluation Metrics}

We evaluate model performance using \texttt{Precision}, \texttt{Recall}, and \texttt{F1-score} with \texttt{Macro} and \texttt{Micro} averaging.
\texttt{Macro} computes scores per class before averaging, while \texttt{Micro} aggregates across all instances.
Both are applied to the classification categories to ensure a balanced assessment. 

\subsection{Model Performance}

\begin{table}[ht!]
  \centering
  \small
  \renewcommand{\arraystretch}{1.2}
  \setlength{\tabcolsep}{6pt}

  \caption{Model overall performance and comparison.}
  \label{tab:overall_results_a}

  \footnotesize
  \setlength{\tabcolsep}{3.5pt}

  \resizebox{\textwidth}{!}{
  \begin{threeparttable}
  \begin{tabular}{lccc|ccc}
      \toprule
      \multicolumn{1}{c}{\multirow{2}{*}{\textbf{Models}}}
          & \multicolumn{3}{c|}{\textbf{Micro-averaged}}
          & \multicolumn{3}{c}{\textbf{Macro-averaged}} \\
      \cmidrule(lr){2-4} \cmidrule(lr){5-7}
          & \textbf{Prec.} & \textbf{Rec.} & \multicolumn{1}{c|}{\textbf{F1}}
          & \textbf{Prec.} & \textbf{Rec.} & \textbf{F1} \\
      \midrule

      \textbf{\modelxcom{}}
          & \textbf{60.51}±0.8 & 56.40±0.7 & \multicolumn{1}{c|}{\textbf{58.38}±0.9}
          & \textbf{61.92}±1.0 & 56.11±0.7 & \textbf{58.56}±0.8 \\

      \midrule
      \multicolumn{7}{l}{\textbf{\colorbox{gray!15}{Feature-based baselines}}} \\
      FastText+SVM
          & 52.57±1.0 & 49.15±0.8 & \multicolumn{1}{c|}{50.08±1.1}
          & 52.83±0.9 & 49.76±0.8 & 51.05±1.0 \\
      FastText+XGBoost
          & 52.03±1.1 & 42.28±1.0 & \multicolumn{1}{c|}{46.65±0.8}
          & 53.65±1.0 & 41.38±0.7 & 46.22±1.1 \\

      \midrule
      \multicolumn{7}{l}{\textbf{\colorbox{gray!15}{Transformer-based baselines}}} \\
      Finetuned-BART
          & 49.70±0.9 & \textbf{61.49}±1.0 & \multicolumn{1}{c|}{54.97±0.9}
          & 50.16±1.1 & \textbf{62.74}±1.0 & 55.72±1.1 \\
      Finetuned-T5
          & \underline{55.88}±0.7 & \underline{57.72}±0.9
          & \multicolumn{1}{c|}{\underline{56.79}±0.8}
          & \underline{55.24}±1.0 & \underline{56.53}±0.9 & \underline{55.86}±0.9 \\

      \midrule
      \multicolumn{7}{l}{\textbf{\colorbox{gray!15}{General-purpose LLM baselines}}} \\
      Llama-3.2-8B-Instruct
          & 24.86±0.9 & 33.90±0.8 & \multicolumn{1}{c|}{28.69±1.0}
          & 25.62±0.7 & 34.69±1.1 & 25.61±0.9 \\
      Gemini-2.5-Flash
          & 46.06±0.8 & 48.96±0.9
          & \multicolumn{1}{c|}{47.47±0.7}
          & 48.97±0.7 & 53.39±0.6 & 44.94±0.8 \\
      ChatGPT-5.1
          & 47.28±0.0 & 50.33±0.0 
          & \multicolumn{1}{c|}{48.76±0.0}
          & 47.61±0.0 & 54.33±0.0 & 47.24±0.0 \\

      \bottomrule
  \end{tabular}

  \begin{tablenotes}
      \small
      \item \textit{Note: Best values in \textbf{bold}, second-best are \underline{underlined}.}
  \end{tablenotes}

  \end{threeparttable}
  }

\end{table}

\begin{table}[ht!]
  \centering
  \small
  \renewcommand{\arraystretch}{1.2}
  \setlength{\tabcolsep}{6pt}

  \caption{Detailed model results of \modelxcom{}.}
  \label{tab:overall_results_b}

  \footnotesize
  \setlength{\tabcolsep}{3.5pt}

  \resizebox{\textwidth}{!}{
  \begin{tabular}{lccccccc}
      \toprule
      \multicolumn{1}{c}{\textbf{\modelxcom{}}}
          & \multicolumn{3}{c|}{\textbf{Micro-averaged}}
          & \multicolumn{3}{c}{\textbf{Macro-averaged}} \\
      \cmidrule(lr){2-4} \cmidrule(lr){5-7}
          & \textbf{Prec.} & \textbf{Rec.} & \multicolumn{1}{c|}{\textbf{F1}}
          & \textbf{Prec.} & \textbf{Rec.} & \textbf{F1} \\
      \midrule

      \multicolumn{7}{l}{\textbf{\colorbox{gray!15}{(A) Per-aspect performance}}} \\
      \addlinespace[3pt]

      \textbf{Appearance}
          & 61.28±0.9 & 60.59±0.8 & \multicolumn{1}{c|}{60.93±0.7}
          & 63.58±1.0 & 59.98±0.9 & 61.18±0.8 \\

      \textbf{Aroma}
          & 60.93±0.8 & 55.27±0.9 & \multicolumn{1}{c|}{57.96±1.1}
          & 61.43±0.9 & 53.57±0.8 & 56.72±0.7 \\

      \textbf{Palate}
          & 56.25±1.0 & 44.75±1.1 & \multicolumn{1}{c|}{49.85±0.9}
          & 64.56±1.1 & 42.79±0.7 & 48.23±0.8 \\

      \textbf{Taste}
          & 61.37±0.7 & 59.73±0.8 & \multicolumn{1}{c|}{60.54±0.6}
          & 61.81±1.0 & 60.71±0.9 & 60.98±0.8 \\

      \midrule

      \multicolumn{7}{l}{\textbf{\colorbox{gray!15}{(B) Component-wise performance}}} \\
      \addlinespace[3pt]

      \textbf{Aspect Classification}
          & 95.38±0.8 & 95.38±0.7 & \multicolumn{1}{c|}{95.38±1.0}
          & 91.69±1.1 & 91.10±0.9 & 91.39±0.8 \\

      \textbf{Comparison Opinion}
          & 61.21±0.9 & 61.21±1.0 & \multicolumn{1}{c|}{61.21±0.8}
          & 62.58±1.1 & 60.91±0.7 & 61.49±0.9 \\

      \textbf{Overall}
          & 60.51±0.8 & 56.40±0.7 & \multicolumn{1}{c|}{58.38±0.9}
          & 61.92±1.0 & 56.11±0.7 & 58.56±0.8 \\

      \bottomrule
  \end{tabular}
  }

\end{table}



Table~\ref{tab:overall_results_a} compares \modelxcom{} with seven baselines using both \texttt{Macro-averaged F1} and \texttt{Micro-averaged F1}, demonstrating \modelxcom{}'s robustness, achieving the highest performance across all metrics (approximately $58.5\%$ \texttt{Macro-averaged F1}).
While feature-based models (SVM, XGBoost), both of which use fastText embeddings as input features, struggle to capture complex contextual dependencies, transformer-based baselines (Finetuned-T5 and BART) prove more competitive.
However, \modelxcom{} outperforms the runner-up (Finetuned-T5) by nearly 3 points, validating the effectiveness of our architecture.
General-purpose LLMs (Llama-3.2-8B-Instruct, Gemini-2.5-Flash, ChatGPT-5.1) show substantially lower performance, with \texttt{Micro-averaged F1} scores ranging from $28\%$ to $49\%$. This reinforces the need for our approach, as fine-grained comparative opinion mining demands structural understanding that zero-shot prompts cannot provide.

Furthermore, \modelxcom{} achieves these results with significantly lower computational cost. Built on BERT-based encoders (approximately $110$M parameters each), it is substantially smaller than multi-billion-parameter LLMs. In our experiments, training \modelxcom{} took around $10$ hours on the same hardware setup, while fine-tuning LLMs typically requires approximately $40$–$80$ hours due to their much larger model size and memory usage. This highlights the computational efficiency of \modelxcom{}, achieving competitive performance with lower resource consumption.

Table~\ref{tab:overall_results_b} presents the performance of the proposed model across four aspects.
\texttt{Appearance} and \texttt{taste} achieve the highest F1 scores exceeding $60\%$, indicating that these aspects contain clearer and more distinctive linguistic cues. Conversely, \texttt{palate} yields the lowest scores due to the subtle and overlapping nature of mouthfeel-related expressions, while \texttt{aroma} shows moderate performance influenced by its lexical proximity to \texttt{taste}.
The component-wise results show that while the Aspect Classification module is highly reliable (surpassing $90\%$ in both \texttt{Macro-averaged F1} and \texttt{Micro-averaged F1}), the subsequent Comparison Opinion extraction drops to around $60\%$ across \texttt{Macro-averaged F1} and \texttt{Micro-averaged F1}. Therefore, the combined effect of these errors contributes to reduced overall results through cascaded error propagation.
These aspect-level and component-level error patterns, along with illustrative examples, are examined in Section~\ref{subsec:error-analysis}.

\subsection{Explanation Results}


\begin{figure}[ht!]
    \centering
    \begin{subfigure}{0.65\textwidth}
        \centering
        \includegraphics[width=\linewidth]{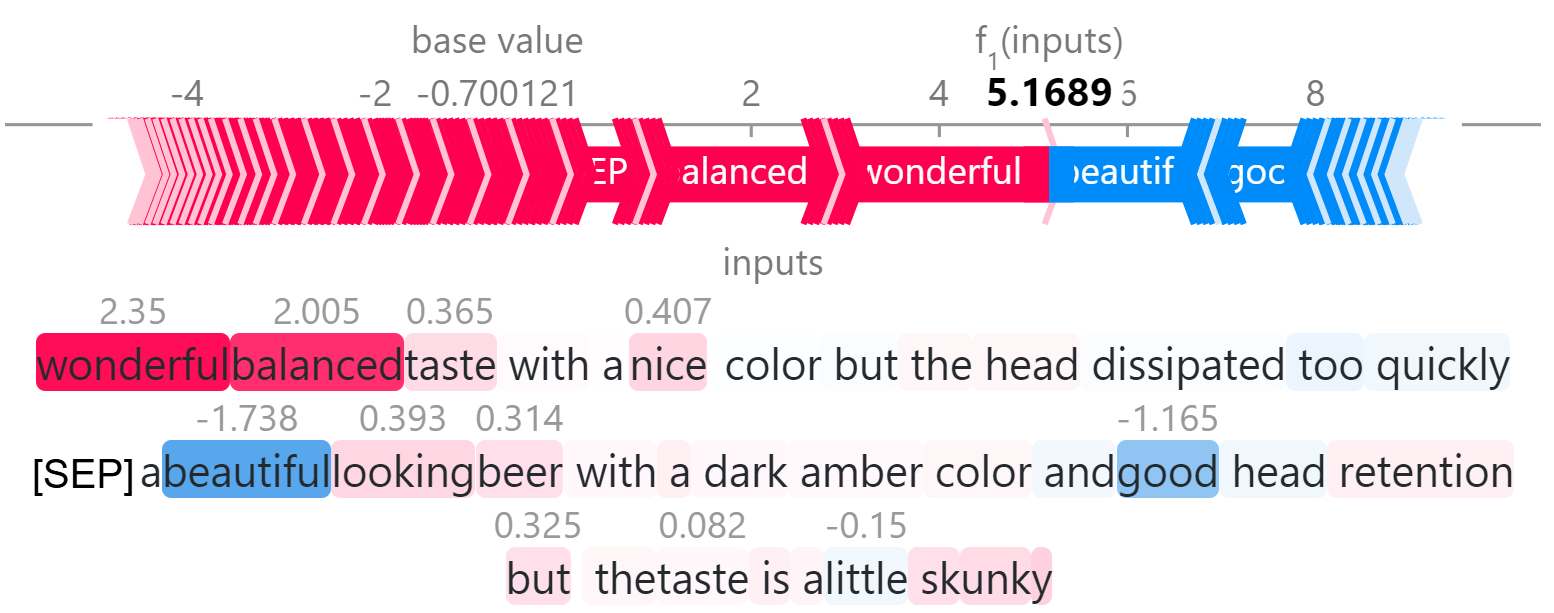}
        \caption{\small Explanation example of the \texttt{taste} aspect comparison.}
        \label{fig:shap-taste}
    \end{subfigure}
    \hfill
    \begin{subfigure}{0.65\textwidth}
        \centering
        \includegraphics[width=\linewidth]{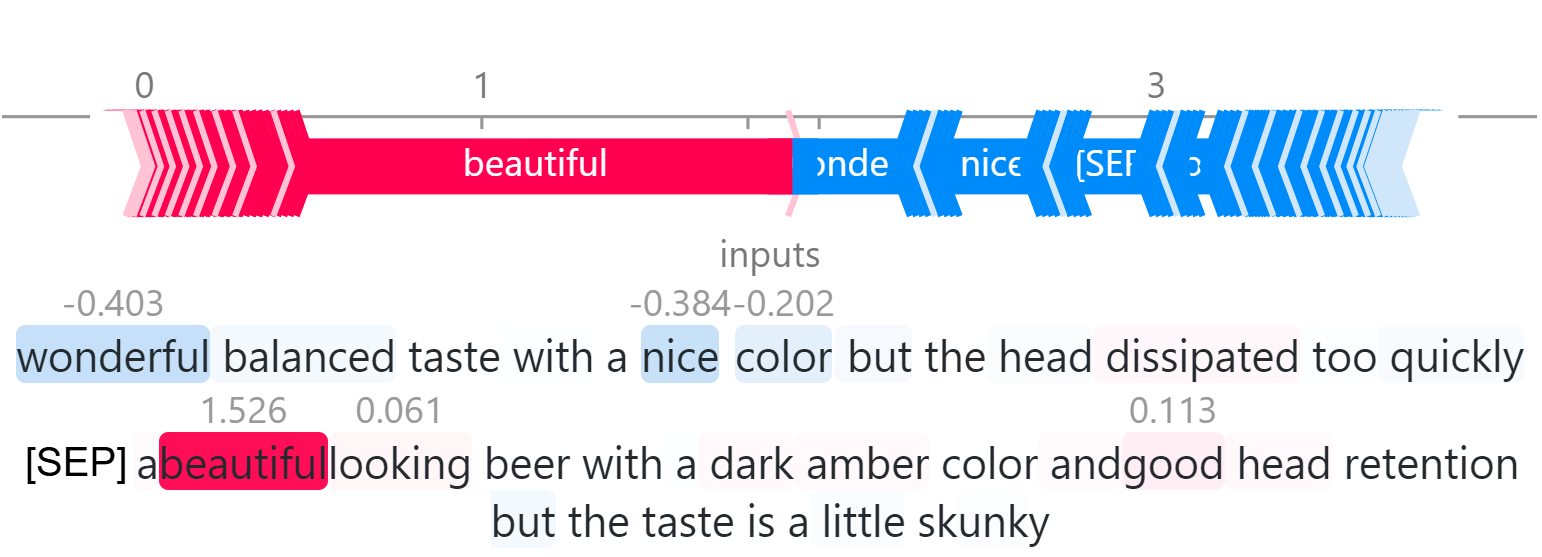}
        \caption{\small Explanation example of the \texttt{appearance} aspect comparison.}
        \label{fig:shap-appearance}
    \end{subfigure}

    \caption{The SHAP-based explanation examples for \texttt{taste} and \texttt{appearance} aspects.
    \footnotesize{
    In our SHAP-based visualization,
    \textcolor[HTML]{FF0051}{red tokens} indicate a positive impact to the model's output, 
    \textcolor[HTML]{008BFB}{blue tokens} indicate negative impact.
    The arrow length represents the absolute SHAP value.
    The following highlighted text provides a more intuitive explanation by marking key tokens in the original text.
    More intense colors signify stronger influence.
    }}
    \label{fig:shap-combined}
\end{figure}

\begin{figure}[ht!]
  \centering
  \includegraphics[width=0.85\columnwidth]{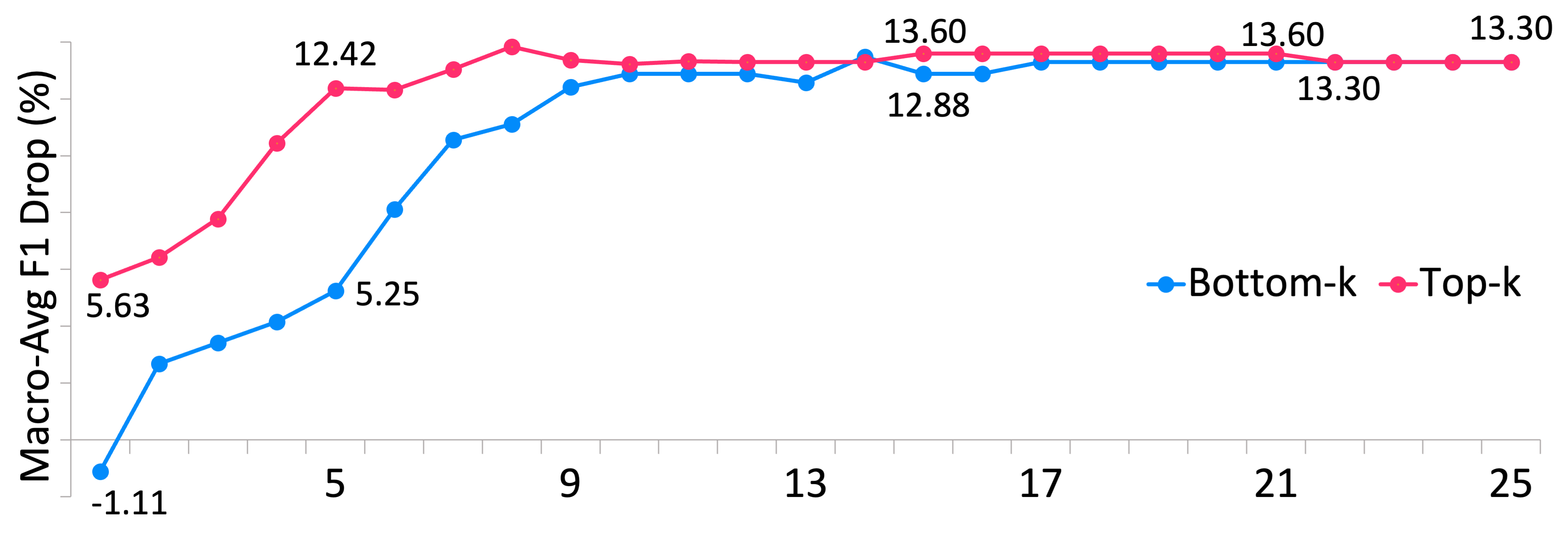}
  \caption{\small SHAP faithfulness evaluation: Macro-F1 drop when removing top-k vs. bottom-k adjectives.}
  \label{fig:xcom_shap_faithfulness}
\end{figure}

Figure~\ref{fig:shap-combined} presents explanations for two aspect-based comparisons (\texttt{Taste} and \texttt{Appearance}) between two reviews of two products. For the \texttt{Taste} aspect (Figure~\ref{fig:shap-taste}), the first product (review 1) is evaluated higher than the second one (review 2). 
Tokens `\textit{wonderful}' and `\textit{balanced}' have a very positive impact on the model output. In contrast, although `\textit{beautiful}' typically conveys a positive sentiment, in the context of the \texttt{Taste} aspect, it has a negative effect since it is not typically used to describe \texttt{taste} and is therefore considered a misleading feature.
With the same input reviews, for the \texttt{appearance} aspect, the first product is evaluated worse than the second product (Figure~\ref{fig:shap-appearance}).
Our model correctly identifies that `\textit{beautiful}' refers to the \texttt{appearance} of the beer, giving it a strong positive impact.
Meanwhile, `\textit{wonderful}' has a weaker influence, reinforcing the idea that the model specializes in its respective aspect when trained on appropriate data.
The sharp contrast between the probability drop when removing top-$k$ and bottom-$k$ tokens supports this finding. 

As shown in Figure~\ref{fig:xcom_shap_faithfulness}, we also evaluated the faithfulness of SHAP by measuring the drop in \texttt{Macro-averaged F1} when removing adjective tokens ranked by importance. The \textcolor[HTML]{FF0051}{Top-k} curve demonstrates an immediate decline in performance, indicating that SHAP correctly identifies the most predictive words, whereas the \textcolor[HTML]{008BFB}{Bottom-k} curve rises slowly, confirming that low-ranked adjectives have minimal impact on the model's decisions. Token importance was determined by averaging the absolute SHAP values across all sentiment classes for tokens in both input sentences. Both curves converge at $k=25$, which represents the maximum number of adjectives found in any single test sample; at this threshold, both strategies result in the removal of all adjectives, leaving identical residual inputs and thus yielding the same performance drop relative to the baseline.

\section{Discussion}
In this section, we present an ablation study to quantify each component’s contribution and an error analysis covering key failure cases--including aspect ambiguity, implicit contrast, and syntactic complexity--to clarify current limitations and guide future improvements.\subsection{Model Component Contribution Analysis}

\begin{figure}[ht!]
    \centering
    \includegraphics[width=0.9\linewidth]{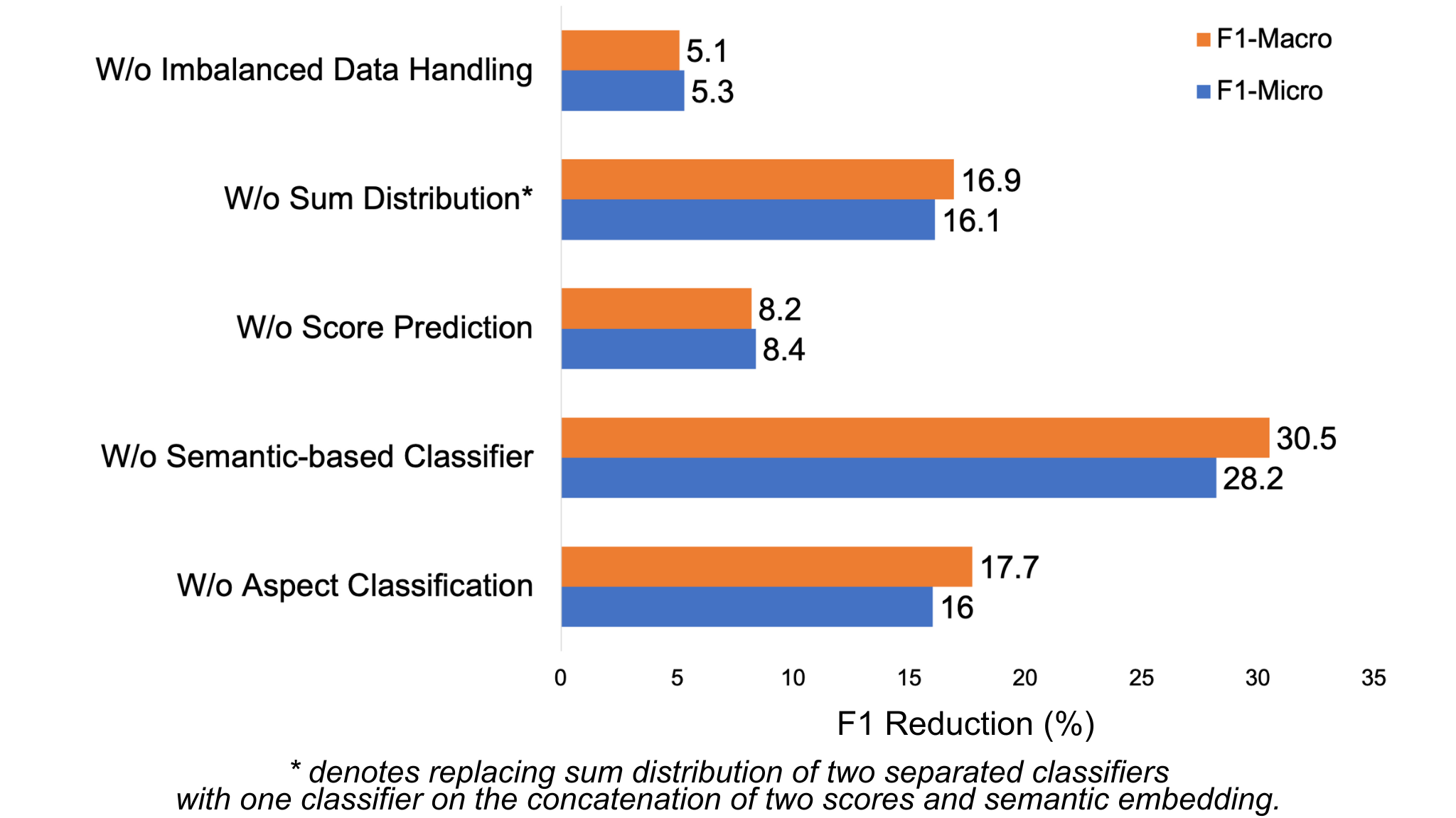}
    \caption{\small Ablation study showing the F1 drop when removing each model component.}
    \label{fig:ablation_test}

\end{figure}

We evaluate each component’s contribution through an ablation study, systematically removing them to measure the performance drop.
Figure~\ref{fig:ablation_test} shows that each component contributes to performance gains, though their impact varies.
The semantic-based classifier is the primary driver for identifying comparative relationships, as its removal leads to a drastic decline of approximately $30\%$ across both \texttt{F1} metrics.
The aspect classification is also indispensable; without it, the model loses its ability to filter relevant opinion targets, leading to a nearly $17\%$ reduction in performance.
Interestingly, replacing the sum distribution strategy with a single classifier over concatenated semantic- and rating-based embeddings results in a performance drop of over $16\%$ -- even greater than removing the entire rating module, suggesting that improper fusion harms representational effectiveness.
Other ablations 
also reduce performance, though more mildly.

\subsection{Error Analysis}
\label{subsec:error-analysis}

\begin{table}[h!]
  \centering
  \small
  \setlength{\tabcolsep}{3pt}
  \renewcommand{\arraystretch}{1.1}
  \caption{\small Error analysis and representative cases of \modelxcom{}.}
  
  \begin{tabular}{p{2.6cm} p{3cm} p{6cm}}
    \toprule
    \textbf{Error type} & \textbf{Cause} & \textbf{Representative examples} \\
    \midrule
    Over-extraction of aspect-related sentence 
      & Model generalizes beyond the intended scope of aspect-related content. 
      & “This beer is very easy to drink” — General statement, not truly about a specific aspect. \\
    \midrule
    Under-extraction of aspect-related sentence 
      & Aspect reference is expressed indirectly or contextually, not explicitly. 
      & “I like to get it from time to time when I want something with style that \textbf{goes down smooth and tasty}” — Missed implicit aspect. \\
    \midrule
    Aspect confusion 
      & A term may refer to multiple aspect categories. 
      & “\textbf{Hops} is in the background however noticeable” — Confused between aroma and taste. \\
    \midrule
    Aspect rating overestimation 
      & Lexical complexity and implicit contrast between co-occurring aspects. 
      & “The \textbf{lace} is amazing and makes this beer almost as good to watch as to \textbf{taste} (well, not quite but you get the idea)” — Model overestimates sentiment for taste. \\
    \midrule
    Incorrect rating prediction 
      & Failure to correctly interpret contrastive expressions. 
      & “It’s sweet but \textbf{not cloying}. In fact, the sweetness comes in and out...” — Model predicts low score, missing contrast from “not cloying”. \\
    \midrule
    Out-of-vocabulary word 
      & The word is missing from the rating dictionary. 
      & “The nose is \textbf{accentuated} with a slight hint of clove” — “Accentuated” absent from dictionary, causing mis-scoring. \\
    \bottomrule
  \end{tabular}

  \label{tab:error_analysis}

\end{table}

Despite its overall effectiveness, the \modelxcom{} model reveals various error types that point to areas needing improvement. Table~\ref{tab:error_analysis} summarizes these errors, their likely causes, and representative examples. A common issue is over-extraction of aspect-related sentences, where the model includes sentences that are generic or not about specific aspects (e.g., “This beer is very easy to drink”). In contrast, under-extraction occurs when aspect-related information is conveyed implicitly rather than explicitly, leading the model to overlook relevant cues. For example, the phrase “goes down smooth and tasty” implies aspects such as palate or taste, yet the model fails to recognize them. Another challenge lies in aspect ambiguity, where a term (e.g., “hops”) can refer to more than one aspect such as taste or aroma. The model also struggles with sentences that express multiple aspects using complex phrasing and subtle contrast, which can lead to overestimation of rating for one aspect due to the strong positivity associated with another. In rating prediction, the model sometimes misinterprets contrastive expressions like “It’s sweet but not cloying,” leading to incorrect rating predictions. Additionally, terms like “accentuated” that are absent from the rating dictionary may receive inaccurate scores due to the model's inability to recognize and interpret their score. These observed errors highlight the need for further improvement, including enhancing contextual understanding, strengthening the model's ability to handle complex syntactic structures, and expanding lexical coverage.

\section{Conclusion}
    

In this paper, we introduced \modelxcom{}, a comparative opinion mining model that takes advantage of several opinion mining modules such as aspect classification, aspect-based rating prediction, and comparative classification. In the first module -- aspect classification, using a specific Transformer-based classifier trained for each aspect, we group sentences from paired reviews into structured sets of aspect-based sentence tuples for comparison. In the second module -- aspect-based rating prediction, there are two key classifiers: a score-based classifier and a semantic-based classifier. In the last module -- comparative classification, both above-mentioned classifiers are to calculate probability distributions. Finally, the model sums these two probability distributions and selects the class with the highest probability as the predicted comparative opinion. 
Furthermore, with SHAP-based explanations, \modelxcom{} enhances transparency beyond conventional black-box models.
The results show that our model yields reliable and interpretable comparative insights, enabling more informed decision-making, while remaining computationally efficient.

\paragraph{Limitations}

Despite the promising results, our approach has several limitations. Below, we discuss these limitations and outline potential directions for future improvements.

Firstly, our model relies on multiple opinion modules (i.e., aspect classification, aspect-based rating prediction, and comparison classifier), which may bring cascading errors if one module propagates incorrect predictions to subsequent stages. Future work could explore a joint learning framework to enable end-to-end optimization.

Secondly, while our SHAP-based explainability module provides insights into key factors influencing the model’s decisions, it is not always easily interpretable for non-expert users. Further research is needed to enhance the readability and usability of explanations, particularly in multi-aspect comparison scenarios.

Thirdly, our SHAP explanations primarily highlight important terms but do not establish a direct comparison between key terms across reviews. While this helps identify influential factors within each review, it does not provide a structured alignment of differing opinions. We are actively exploring a graph-based explanation approach that directly connects key terms from both reviews, intending to provide a more structured and comparative analysis.

Fourthly, the SUDO dataset is limited in scale and domain diversity. 
While SUDO provides a focused evaluation of implicit comparative opinion mining in the beer review domain, validating generalizability requires extending to other product categories. Extending experiments and datasets to include a wider variety of domains and user perspectives would improve the generalizability of our findings.

Fifthly, our data survey reveals that user reviews on different aspects often exhibit interdependencies. Specifically, users who rate one aspect positively are likely to rate other aspects similarly or display a positive bias. We plan to explore cross-aspect dependencies to improve model performance and enhance the explanation component by incorporating these interrelationships.

Finally, our model lacks a specialized mechanism to detect contradictory or misleading reviews caused by informal writing styles or slang, which is a common challenge in text processing for social domains. Future work could explore incorporating linguistic normalization techniques to standardize informal language and slang before analysis. On the other hand, leveraging a large-scale pre-trained model on social media text may improve the model's robustness to informal language patterns commonly found in user-generated content.

By addressing these limitations, we believe that future research can further enhance the effectiveness, interpretability, and applicability of \modelxcom{} in real-world comparative opinion mining tasks.

%
%

\bibliographystyle{splncs04}
\bibliography{custom.bib}

\end{document}